%% file: root.tex
\documentclass[letterpaper, 10 pt, conference]{ieeeconf}  

\IEEEoverridecommandlockouts

\usepackage{graphicx} 
\usepackage{amsmath} 
\usepackage{amssymb}  
\usepackage{xcolor}
\usepackage{xspace}
\usepackage{url}

\usepackage{array}
\usepackage{booktabs}

\usepackage{hyperref}
\usepackage{subfigure}
\usepackage{bytefield}
\usepackage{multirow}
\usepackage{tikz}
\usepackage{rotating}
\usetikzlibrary{positioning, shapes, arrows, automata}
\usepackage{glossaries}
\glsdisablehyper 
\usepackage[normalem]{ulem}

\usepackage{marginnote}

\hypersetup{
    colorlinks,
    linkcolor={red!50!black},
    citecolor={blue!50!black},
    urlcolor={blue!80!black}
}

\newif\ifdraftcolor

\ifdraftcolor
\newcommand{\fer}[1]{{\color{orange}#1}}
\newcommand{\todo}[1]{{\color{red}#1}}
\newcommand{\zac}[1]{{\color{green}#1}}

\newcommand{\authorlist}{Authors}
\else
\newcommand{\fer}[1]{#1}
\newcommand{\todo}[1]{#1}
\newcommand{\zac}[1]{#1}

\newcommand{\authorlist}{Fernando Cladera, Zachary Ravichandran, Ian D. Miller, M. Ani Hsieh, C. J. Taylor, and Vijay Kumar}
\fi

\newacronym{cots}{COTS}{commercial-off-the-shelf}
\newacronym{manet}{MANET}{mobile ad-hoc mesh network}
\newacronym{uav}{UAV}{unmanned aerial vehicle}
\newacronym{phy}{PHY}{physical layer}
\newacronym{snr}{SNR}{signal-to-noise ratio}
\newacronym{rtt}{RTT}{round-trip time}
\newacronym{rssi}{RSSI}{received signal strength indicator}
\newacronym{ugv}{UGV}{unmanned ground vehicle}

\title{\LARGE \bf
Enabling Large-scale Heterogeneous Collaboration with\\ Opportunistic Communications
}

\newcommand{\papertitle}{MOCHA\xspace}

\author{\authorlist%
\thanks{All authors are with GRASP Laboratory, University of Pennsylvania.
Corresponding author: {\tt\small fclad@seas.upenn.edu}.}
\thanks{
We gratefully acknowledge the support of
ARL DCIST CRA W911NF-17-2-0181, 
NSF Grants CCR-2112665, 
ONR grant N00014-20-1-2822, 
ONR grant N00014-20-S-B001, 
NVIDIA, 
and C-BRIC, a Semiconductor Research Corporation Joint University Microelectronics Program program cosponsored by DARPA. 
}
}

\begin{document}

\maketitle
\thispagestyle{empty}
\pagestyle{empty}

\begin{abstract}
\input{sections/abstract}
\end{abstract}

 The source code for \papertitle and the high-altitude UAV planning system is available open source\footnote{\url{http://github.com/KumarRobotics/MOCHA}\\\url{http://github.com/KumarRobotics/air_router}}.


\glsresetall

\input{sections/introduction}
\input{sections/related_work}
\input{sections/methods}
\input{sections/experiments}
\input{sections/conclusion}

\addtolength{\textheight}{-6cm}   


\section*{ACKNOWLEDGMENT}
The authors would like to acknowledge Alex Zhou for helping assemble and service our \glspl{uav}. We would like to thank Kumar Robotics team members, Alice Kate Li and Priyanka Shah, for their help during field experiments. Finally, we would like to thank Dr. Ethan Stump from ARL and Dr. Barbara Dallap Schaer from Penn Vet for their collaboration with test locations for the field experiments.

\bibliographystyle{IEEEtran}
\bibliography{IEEEabrv, literature}

\end{document}

%% file: sections/abstract.tex
Multi-robot collaboration in large-scale environments with limited-sized teams and without external infrastructure is challenging, since the software framework required to support complex tasks must be robust to unreliable and intermittent communication links. In this work, we present \emph{\papertitle} (Multi-robot Opportunistic Communication for Heterogeneous Collaboration), a framework for resilient multi-robot collaboration that enables large-scale exploration in the absence of continuous communications. \papertitle is based on a gossip communication protocol that allows robots to interact opportunistically whenever communication links are available, propagating information on a peer-to-peer basis. We demonstrate the performance of \papertitle through real-world experiments with \gls{cots} communication hardware. We further explore the system's scalability in simulation, evaluating the performance of our approach as the number of robots increases and communication ranges vary. Finally, we demonstrate how \papertitle can be tightly integrated with the planning stack of autonomous robots. We show a communication-aware planning algorithm for a high-altitude aerial robot executing a collaborative task while maximizing the amount of information shared with ground robots.

%% file: sections/introduction.tex
\begin{figure}[t]
    \centering
    \includegraphics[width=\linewidth]{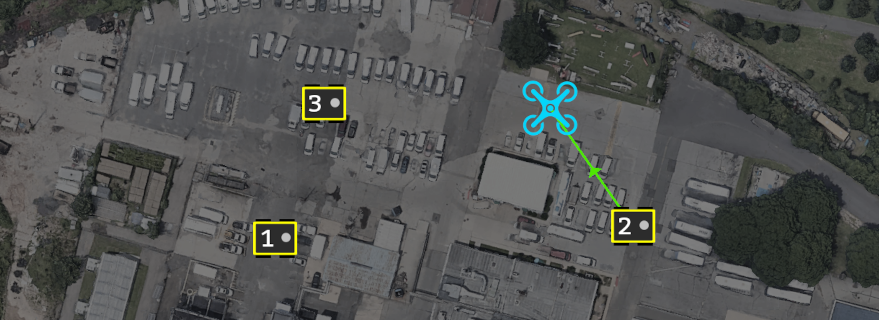}

    \vspace{.1cm}
    \includegraphics[width=\linewidth]{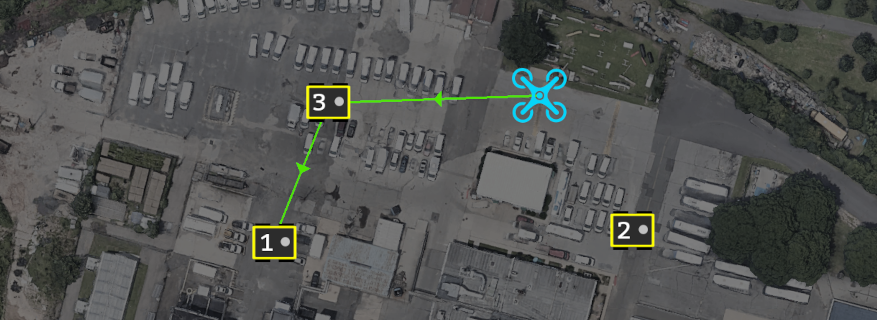}

    \vspace{.1cm}
    \includegraphics[width=\linewidth]{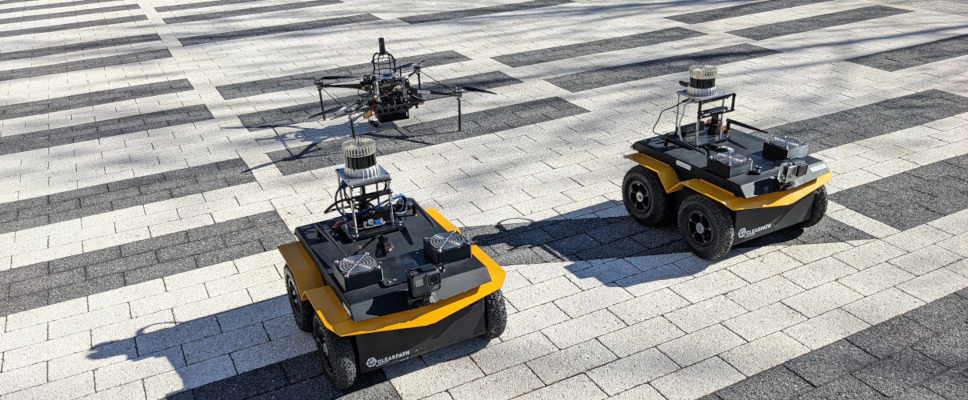}
    \caption{\uline{Top}: Visualization of the real-world experiments using \papertitle to enable communications between a UAV and three UGVs. The UAV first communicates with robot 2 and exchanges information. \uline{Middle}: The UAV acts as a data mule transmitting robot 2's information to robot 3, which itself propagates the information to robot 1. \uline{Bottom}: Autonomous robots during our heterogeneous exploration experiments: the Falcon 4 UAV and Clearpath Jackals.}
    \label{fig:fig1}
    \vspace{-.8cm}
\end{figure}

\section{Introduction}
Air-ground collaboration has proven to be an effective approach to performing autonomous large-scale exploration of unknown environments~\cite{shen2017collaborative, miller2022stronger, liu2022review}.
\Glspl{ugv}  provide stable platforms for large sensors, and can safely approach and inspect objects of interest on the ground. However, the sensors' field of view and range can be limited by terrestrial obstacles like trees and the \gls{ugv}'s motion is constrained by traversability limitations and speed considerations.

\Glspl{uav}, by contrast, are not confined to two-dimensional motion, can avoid many obstacles by flying at higher altitudes, and can provide
vantage points that cover more of the scene.
However, \glspl{uav} expend significantly more energy on locomotion~\cite{mulgaonkar2019small}. Therefore, heterogeneous robot teams can harness the advantages of both air and ground vehicles when exploring large-scale environments to outperform either modality alone. 

In this context of robot collaboration, a reliable communication system is critical for successful mission completion. 
Communication is necessary for coordinating goals, sharing maps, and informing operators about the progress of a mission~\cite{miller2022stronger, tranzatto2022cerberus, otsu2020supervised}.
Existing works focus on addressing the challenges of data fusion~\cite{durrant2016multisensor}, development of robust consensus algorithms~\cite{tahbaz2008consensus, saulnier2017flocking}, efficient task allocation \cite{malencia2021fair}, and probabilistic and graph-theoretic approaches for achieving collaborative behavior~\cite{edwards2022stochastic, saldana2017resilient}. 
However, most frameworks do not address the specific needs of multi-robot collaborative missions when guarantees on communication capability cannot be provided.
Instead, they attempt to try to maintain connectivity~\cite{mox2022learning} or establish links in a predictable fashion at desired rendezvous points~\cite{hsieh2008maintaining, kantaros2018distributed, xi2020synthesis}. 
Nevertheless, when team sizes are limited and the physical scale of the environment is large, it makes sense to allow vehicles to move out of each other's communication range in order to maximize the area explored by the team in a finite amount of time.
As such, we would expect individual vehicles to be operating in isolation or forming small communication cliques for most of the task execution time in real-world settings.

In this work, we present \papertitle,
a communication and coordination framework for collaboration in large-scale environments with a limited number of robots. 
We assume that the individual communication range of each robot is a small fraction of the environment's total size. In our gossip-based framework, two or more robots exchange information whenever they are in communication range.  Gossip approaches are  \emph{convergently consistent}, are simple to implement and analyze, are robust to network disruptions~\cite{birman2007gossip}, and are particularly well-suited for low-volume exchanges such as those in robotics applications with relatively small teams. \papertitle provides most of the advantages of consensus algorithms~\cite{aditya2021survey} for communication without the computational complexity. Moreover, as the communication between robots is negotiated, \papertitle incurs less channel contention than flooding protocols~\cite{mezei2010robot}.

We provide a self-contained and complete system for real-world robot team communication, focusing on providing a usable software stack to enable further multi-robot research in the field using \gls{cots} communication hardware.
\papertitle has been extensively tested in multiple multi-robot deployments; a preliminary version was used for our DARPA SubT deployments~\cite{miller2020mine}, and we have improved it for over two years before public release.

To sum up, \textbf{our contributions with \papertitle are}:
\begin{itemize}
    \item An open-source collaboration framework for robots operating with intermittent communications. Existing and future coordination algorithms can be incorporated into our framework for large-scale field deployments of heterogeneous robot teams. 
    \item Extensive large-scale real-world experiments to validate our system with five robots. In addition, we simulate \papertitle to evaluate system performance with increased numbers of robots over varying communication ranges.
    \item An open-source planner for a high-altitude \gls{uav} that demonstrates the utility of using \papertitle to inform autonomy for actively optimizing communications.
\end{itemize}

%% file: sections/related_work.tex
\section{Related Work}

Communication in field robotics has been a longstanding problem for multi-robot field experiments. It can be separated into two paradigms: 1) ensured network connectivity and 2) opportunistic communication. We will provide a brief overview of these methods in this section.

\subsection{Ensured Network Connectivity}
Ensured network connectivity approaches rely on having enough agents to cover a particular area, deploying relay nodes as needed, and constructing a mesh network where packets can be routed between the nodes. Ad-hoc networks with the proper routing protocols~\cite{chroboczek2011babel} or \Gls{cots} solutions, such as \glspl{manet}, can generate appropriate mesh networks. For instance, in~\cite{hsieh2008maintaining}, robots can perform a mission while keeping connectivity through reactive controllers.
Other works focus on deploying robots in specific configurations~\cite{mox2022learning} to guarantee continuous communication~\cite{stump2011visibility}.

Ensured network connectivity systems are advantageous when the nodes require real-time feedback from an operator to provide high-level commands. For instance, some teams at the DARPA SubT challenge deployed \glspl{manet} nodes, generated mesh networks to extend communications among all nodes, or used wired backbones to extend communication~\cite{tranzatto2022cerberus}. 

\subsection{Opportunistic Communication}
Opportunistic strategies do not aim to maintain connectivity, but rather take advantage of it when available.
Opportunistic communication offers significant advantages over continuous communication approaches: by relaxing the fully-connected constraint, the number of nodes in the network can be reduced, and mobile agents can act as \emph{data mules}~\cite{cheng2018air}, propagating messages far beyond the communication range of the wireless transceivers.

Recently, DDS-based~\cite{dds} approaches based in  ROS 2~\cite{macenski2022robot} have been used for communication in multi-robot settings, such as CHORD~\cite{Ginting2021CHORDDD}. However, ROS 2/DDS approaches based on the \emph{reliable} QoS are only suitable for short intermittent disconnections, as networks become congested when a node reconnects after an extended period of time.
The authors aimed to improve congestion management when nodes reconnect after a long period of inactivity in ACHORD~\cite{saboia2022achord}. They did this by modeling communication channels and using communications-aware coordination based on the size of message queues. Still, there is a limit to how long robots can explore since they need to get within communication range periodically to flush their queues.

{Compared to existing solutions, \papertitle is completely opportunistic. It does not require periodic rendezvous, radio deployments, or continuous communications between robots. Our communication approach is robust and predictable and reduces network congestion by cleverly propagating only the latest information available from each node. Similar to~\cite{saboia2022achord}, we provide metrics to inform higher-level planning decisions in the nodes. Our framework, however, does not impose restrictions on how long the nodes can explore without communication. }

%% file: sections/methods.tex
\begin{figure}[t]
    \centering
    \begin{bytefield}[bitwidth=.5em]{48}
        \bitheader{0, 8, 16, 32, 48} \\
        \bitbox{8}{RID}
        & \bitbox{8}{TID}
        & \bitbox{16}{Time [s]}
        & \bitbox{16}{Time [ms]}
    \end{bytefield}
    \caption{Structure of a header. RID stores the robot ID, whereas TID stores the topic ID.
    Our communication system can manage a maximum of 256 nodes, each with up to 256 topics.
    }
    \label{fig:header}
    \vspace{-.5cm}
\end{figure}

\begin{figure*}[t]
    \centering
    \includegraphics[width=.68\linewidth]{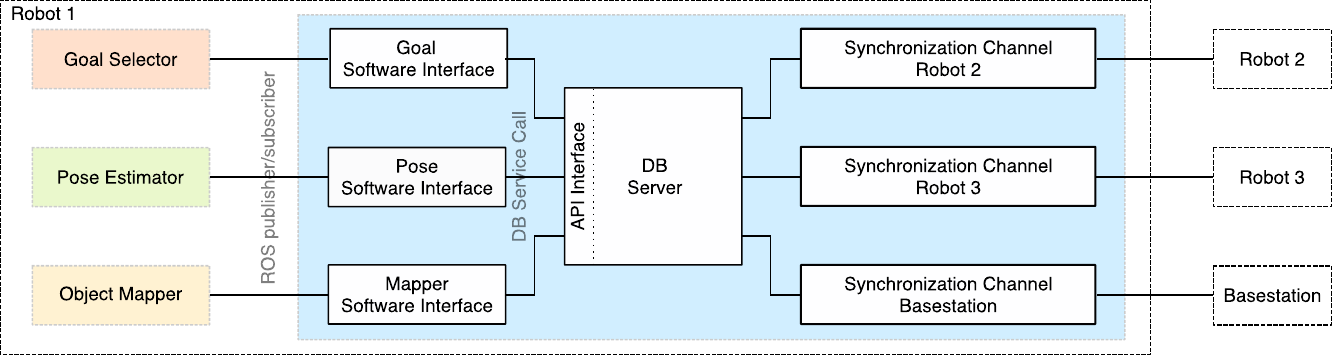}
    \vspace{-.3cm}
    \caption{Communication stack architecture from a single robot point of view.
    A shared configuration file defines the topics committed to the database and
    network topology, which generates the required software interfaces and synchronization channels.}
    \label{fig:commStackArchitecture}
    \vspace{-.5cm}
\end{figure*}

\section{Methods}
\label{sec:methods}
This section describes the design philosophies behind \papertitle and the details of the metrics provided by the system.

\subsection{Distributed Communication System}

The architecture of the communication stack for each robot in the team is shown in Fig.~\ref{fig:commStackArchitecture}.  
The communication stack is composed of the following elements:
\begin{itemize}
    \item \textbf{Database (DB) Server}: a key-value storage that records all
      the system's messages. Robots can store messages produced locally
      through software interfaces or get messages from other robots through a synchronization channel. The DB provides a simple API to store and retrieve messages.
    To reduce network traffic, all payloads are throttled and compressed with LZ4.  It is worth noting that data from each robot is stored separately, allowing robots to store-and-forward other agents' data.
    \item \textbf{Software Interfaces}: These interface the rest of the robot
      middleware with \papertitle. For instance, when using ROS, the software
      interfaces subscribe to other ROS topics and perform the required DB API calls to insert the data into the DB Server. 
    Conversely, when the DB server receives messages from another robot, the
    software interfaces publish the messages in the namespaced ROS topics. 
    \item \textbf{Synchronization channels}: perform the peer-to-peer synchronization between two nodes.
    The synchronization channels are based on ZeroMQ~\cite{hintjens2013zeromq}, following the request/reply model. We have experimentally proven that this approach provides the highest reliability for intermittent communication. The synchronization process is described in detail in Sec.~\ref{subsubsec:sync}.
\end{itemize}

\subsubsection{Message Headers}
\papertitle relies on a key-value storage model. All messages are stored in an
in-memory database on each robot, where the 
key is the message header, and the value is the binary payload of the message.
An example of a message header is shown in Fig.~\ref{fig:header}. Headers are computed when a robot inserts a message into its own database.

A header is composed of the following elements:
\begin{itemize}
    \item \textbf{RID} and \textbf{TID}: These fields identify the robot and topic ID of the message,
    respectively.
    \item \textbf{Time [s]} and \textbf{Time [ms]}: store the local robot time that produced the message.
\end{itemize}

\papertitle relies on a configuration file shared among all nodes that describes the different nodes in the network, the topology, and which topics are published by each robot. RID and TID are computed from the configuration file.

\subsubsection{Synchronization}
\label{subsubsec:sync}
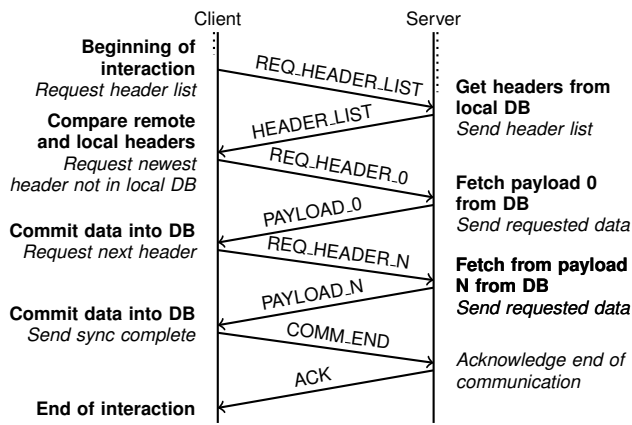
\begin{figure}[t]
    \input{media/exchange}
    \vspace{-.5cm}
    \caption{Synchronization procedure for the distributed communication system. Any node in the system can perform the client or server role. 
    Bold text represents the action of \papertitle in the robot, whereas the text in italics represents the message transmitted to the other peer.
    We consider a complete interaction when the client reaches the \textbf{End of interaction} stage.}
    \label{fig:synchro}
    \vspace{-.5cm}
\end{figure}

The synchronization process is shown in Fig.~\ref{fig:synchro}. For simplicity, we will consider the point of view of a \emph{client} robot fetching information from a remote \emph{server}. Any node in the system can perform the client or remote server role, and in general, perform both roles in different exchanges. 

The client starts by requesting the lists of headers from the server, which is a concatenated stream of all its most recent headers. The header list size is $S_h = \sum_i r_i$, where $r_i$ is the number of topics provided by robot $i$ on the network.
As the size of the header is only 6 bytes (Fig.~\ref{fig:header}), the exchange of all the headers is a fast operation that usually happens within a single ZeroMQ frame\footnote{\url{https://rfc.zeromq.org/spec/13/}}. This is crucial to keep the amount of information exchanged bounded: a client $i$ can request at most $S_h - r_i$ messages in the worst-case scenario. 

The client robot compares its local list of headers with the one from the remote server and decides which messages to request based on the timestamp of the messages. Messages are requested in order based on their predefined priority.
Payload transmissions are performed individually through synchronization channels. The client commits the message into its database once a transmission from the server is successful.

The client does not immediately discard old messages for a particular RID and TID to remove potential race or lock conditions during the communication exchange. This scenario occurs when three robots form a communication chain: $A \rightarrow B \rightarrow C$.
In this case, C may request information from A through B, while B may be updating messages from A directly.

In case of disconnection, there are two possible scenarios:
\begin{itemize}
    \item Short-term disconnection: ZeroMQ can recover from intermittent disconnection, and a message may continue transmission. A polled timeout of 5 seconds was set for this condition in our experiments. 
    \item Long-term disconnection: the synchronization fails, the interaction is interrupted and it will be performed again from the beginning if opportunity allows.
\end{itemize}

\subsubsection{Additional Features}
\papertitle requires no explicit clock synchronization between nodes. All the header data processing on the client side compares only previous messages from the same robot.

The synchronization procedure can be triggered based on information from the \gls{phy}, such as a predefined \gls{snr} or \gls{rssi} threshold. Alternatively, a recurrent synchronization timer can be set for every communication channel.

Finally, while our system does not explicitly support sending incremental updates of topics (such as position updates), the system will aggregate small messages into larger database updates which reduces network overhead substantially and allows the synchronization complexity to scale linearly with the number of topics rather than the number of messages.


\subsection{Network and Communication Metrics}
\label{subsec:metrics}
\papertitle provides high-level information about the communication status between nodes. These include:
\begin{itemize}
    \item Last synchronization time with a particular node.
    \item Link quality, measured as \gls{rtt} and network bandwidth. Metrics from the network physical layer can also be provided, such as \gls{snr} or \gls{rssi}.
    \item Status of the exchange: idle, comm start, comm end, timeout.
\end{itemize}
These metrics can be used to inform high-level planning algorithms, as shown in Sec.~\ref{subsec:comm-aware-expl}.

%% file: media/exchange.tex
\begin{tikzpicture}[font=\sffamily,thick,
  commentl/.style={text width=2.5cm, align=right},
  commentr/.style={commentl, align=left},]
  \scriptsize

  \node[] (init) {Client};
  \node[right=2cm of init] (recv) {Server};

  \draw[->] ([yshift=-0.5cm]init.south) coordinate (reqheader1o) --
    ([yshift=-.5cm]reqheader1o-|recv) coordinate (reqheader1e)
    node[pos=.55, above, sloped] {REQ\_HEADER\_LIST};

  \draw[->] ([yshift=-.1cm]reqheader1e) coordinate (headerlist1o) --
    ([yshift=-.5cm]headerlist1o-|init) coordinate (headerlist1e)
    node[pos=.55, above, sloped] {HEADER\_LIST};

  \draw[->] ([yshift=-.1cm]headerlist1e) coordinate (getheader1o) --
    ([yshift=-.5cm]getheader1o-|recv) coordinate (getheader1e)
    node[pos=.55, above, sloped] {REQ\_HEADER\_0};

  \draw[->] ([yshift=-.1cm]getheader1e) coordinate (getdata1o) --
    ([yshift=-.5cm]getdata1o-|init) coordinate (getdata1e)
    node[pos=.55, above, sloped] {PAYLOAD\_0};

  \draw[->] ([yshift=-.1cm]getdata1e) coordinate (getheadern1o) --
    ([yshift=-0.5cm]getdata1e-|recv) coordinate (getheadern1e)
    node[pos=.55, above, sloped] {REQ\_HEADER\_N};

  \draw[->] ([yshift=-.1cm]getheadern1e) coordinate (getdatan1o) --
    ([yshift=-.5cm]getdatan1o-|init) coordinate (getdatan1e)
    node[pos=.55, above, sloped] {PAYLOAD\_N};

  \draw[->] ([yshift=-.1cm]getdatan1e) coordinate (getend1o) --
    ([yshift=-0.5cm]getdatan1e-|recv) coordinate (getend1e)
    node[pos=.55, above, sloped] {COMM\_END};

  \draw[->] ([yshift=-.1cm]getend1e) coordinate (getdataend1o) --
    ([yshift=-.5cm]getdataend1o-|init) coordinate (getdataend1e)
    node[pos=.55, above, sloped] {ACK};



  \draw[thick, shorten >=-.2cm] (init) -- (init|-getdataend1e);
  \draw[thick, shorten >=-.2cm] (recv) -- (recv|-getdataend1e);

  \draw[dotted] (recv.285)--([yshift=2mm]recv.285|-reqheader1e)
  coordinate[pos=.5] (aux1);

  \draw[dotted] (init.255)--([yshift=2mm]init.255|-reqheader1o);



  \node[left = 2mm of reqheader1o.west, commentl]
  {\textbf{Beginning of interaction}\\{\itshape Request header list}};
  \node[left = 2mm of headerlist1e.west, commentl]
  {\textbf{Compare remote and local headers}\\
  {\itshape Request newest header not in local DB}};
  \node[left = 2mm of getdata1e.west, commentl]
  {\textbf{Commit data into DB}\\
  {\itshape Request next header}};
  \node[left = 2mm of getdatan1e.west, commentl]
  {\textbf{Commit data into DB}\\{\textit{Send sync complete}}};
    \node[left = 2mm of getdataend1e.west, commentl]
  {\textbf{End of interaction}};
  

  \node[right = 2mm of reqheader1e.west, commentr]
  {\textbf{Get headers from local DB}\\
  {\itshape Send header list}};
  \node[right = 2mm of getdata1o.west, commentr]
  {\textbf{Fetch payload 0 from DB}\\
  {\itshape Send requested data}};
  \node[right = 2mm of getdatan1o.west, commentr]
  {\textbf{Fetch from payload N from DB}\\
  {\itshape Send requested data}};
  \node[right = 2mm of getdatan1o.west, commentr]
  {\textbf{Fetch from payload N from DB}\\
  {\itshape Send requested data}};
    \node[right = 2mm of getdataend1o.west, commentr]
  {\textit{Acknowledge end of communication}};
\end{tikzpicture}

%% file: sections/experiments.tex
\section{Experiments}
\label{sec:experiments}

\subsection{Communication-aware Exploration}
\label{subsec:comm-aware-expl}
We leveraged the metrics provided by \papertitle to inform the planning stack for a high-altitude \gls{uav}. This experiment builds on our previous work~\cite{miller2022stronger}: the \gls{uav} performs an exploration task in an unknown environment by following a set of predefined waypoints. While exploring the environment, the aerial robot generates a semantic map transmitted to multiple ground robots in the field. Ground robots use this map to localize themselves, find objects of interest (such as cars), and plan trajectories toward these objects of interest, which we refer to as goals.

\todo{We observed that when the \gls{uav} performs only exploration and passive communication, increasing the number of ground robots resulted in diminishing returns as maps are not transmitted back to ground robots regularly. 
The communication-aware exploration proposed in this section addresses the shortcomings of our previous work.}

\begin{figure}
    \centering
    \input{media/sm}
    \vspace{-.5cm}
    \caption{State machine describing the communication-aware exploration mission for the UAV. Transitions between states occur when timers elapse (timeout) or when an event occurs, such as finding a ground robot or ending communications. In our experiments, we used $t_i = 3\,\text{min}$, $t_e = 2\,\text{min}$, $t_s = 45\,\text{s}$, and $t_c = 20\,\text{s}$.}
    \label{fig:uav_sm}
    \vspace{-.5cm}
\end{figure}
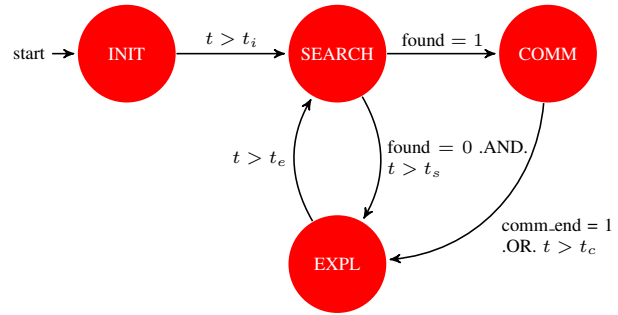

In our strategy, the \gls{uav} transitions between two main modes: exploration and communication. The state machine describing the algorithm is shown in Fig.~\ref{fig:uav_sm}. The \gls{uav} starts its mission with an initial exploration task (INIT), following a predefined set of waypoints. After a predetermined time, it transitions to a search mode (SEARCH), where it seeks one of the ground robots to communicate in a round-robin fashion. This search is performed using the same waypoints as the exploration task,  planning new routes between them, and avoiding no-fly zones. The last known position of the ground robot or the target position is used as a heuristic for where to search. If the ground robot is found, the \gls{uav} acquires the current position of the ground robot and proceeds to transition to COMM, where it flies over the new robot position until all communications are finished. Otherwise, the \gls{uav} transitions into a new exploration task (EXPL), where it continues to explore waypoints.
If the exploration mission finishes (all waypoints are reached), the \gls{uav} only performs search missions.

\begin{figure}
    \centering
    \includegraphics[width=.75\linewidth]{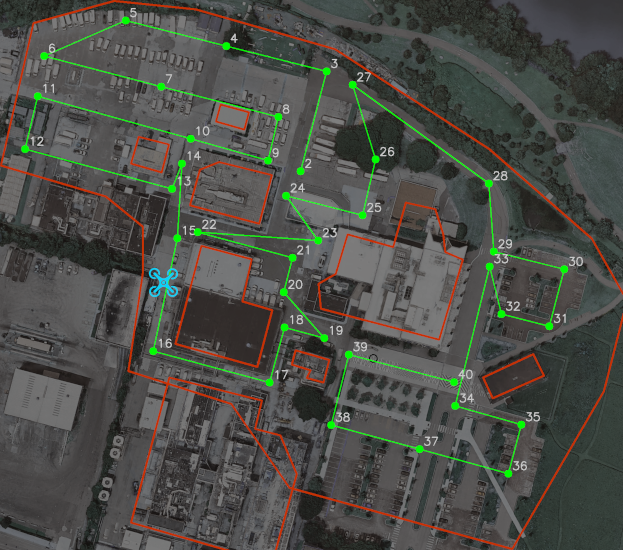}
    
    \vspace{.1cm}    \includegraphics[width=.75\linewidth]{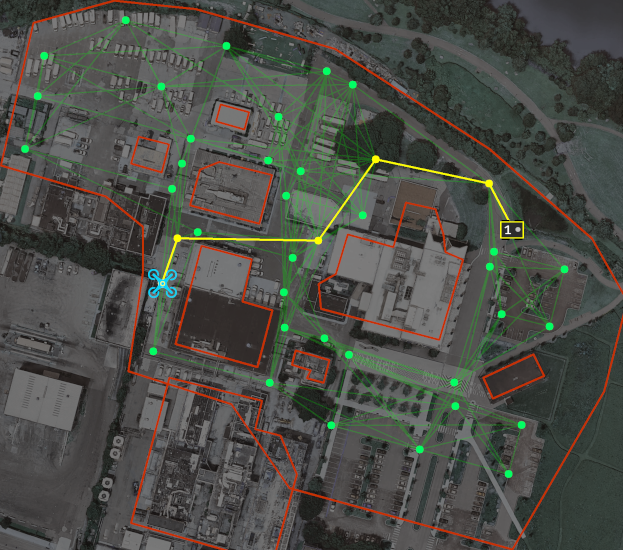}
    \caption{Communication-aware planning. \uline{Top}: A high-altitude UAV performs an {exploration} mission on a set of predefined waypoints and transmits the aerial map to ground robots deployed in the field (INIT and EXPL states). \uline{Bottom}: The UAV transitions to a search mission (SEARCH state), searching for UGV 1 using predefined routes. When a ground robot is found, the \gls{uav} transitions into COMM state and flies over the ground robot's position to maximize communication bandwidth. Red lines represent the no-fly zones.}
    \label{fig:air_router}
    \vspace{-.3cm}
\end{figure}

We defined the \gls{uav} waypoints manually for safety reasons, but alternatively, these can be generated automatically with a coverage algorithm. The \gls{uav} performs passive communication during EXPL and INIT states too, but the quality of the channel deteriorates rapidly as the \gls{uav} flies away from the ground robot.


\begin{table}[t]
    \centering
    \footnotesize
    \noindent
    \input{tables/simresults}
    \caption{Simulation results for multi-robot experiments. }
    \label{tab:results_sim}
    \vspace{-.5cm}
\end{table}

\begin{figure}[t]
    \centering
    \includegraphics[trim={3.2cm 0cm 3.2cm 1cm},clip, width=\linewidth]{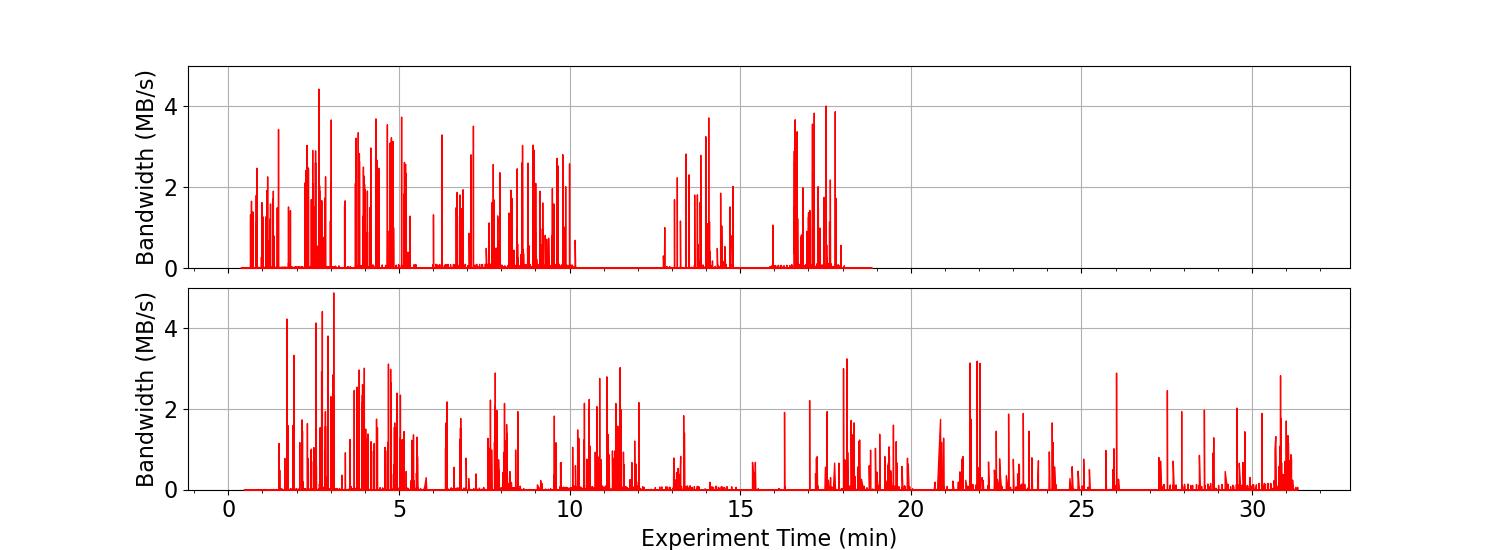}
    \caption{Total bandwidth evolution for two real-world experiments with one \glspl{uav} and three \glspl{ugv}.  We can observe that communications are sporadic, with extended silent periods when robots are not within communication range.  }
    \label{fig:real-world-bw}
    \vspace{-.7cm}
\end{figure}

\begin{figure*}
    \centering
    \begin{sideways}
    \begin{minipage}{.22\linewidth}
    \centering
            \textbf{Real-world}
    \end{minipage}
    \end{sideways} \,
    \includegraphics[height=.22\linewidth]{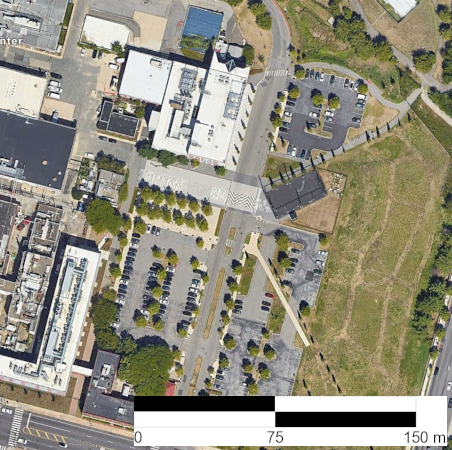}
    \includegraphics[height=.22\linewidth]{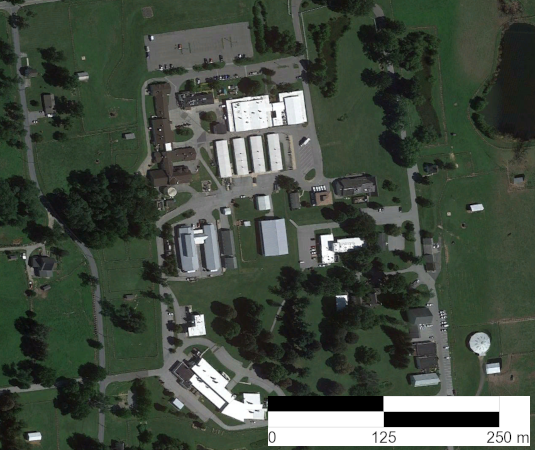}
    \includegraphics[height=.22\linewidth]{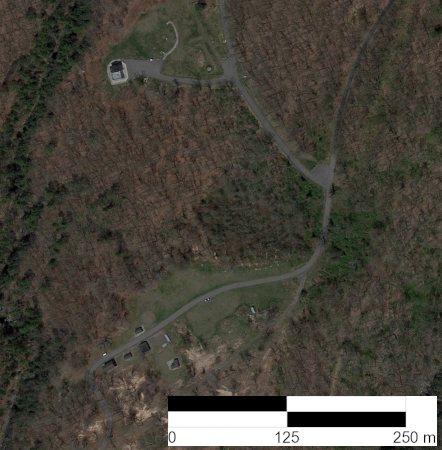}

    \vspace{.1cm}
    \begin{sideways}
    \begin{minipage}{.22\linewidth}
    \centering
            \textbf{Simulation}
    \end{minipage}
    \end{sideways} \,
    \includegraphics[height=.22\linewidth]{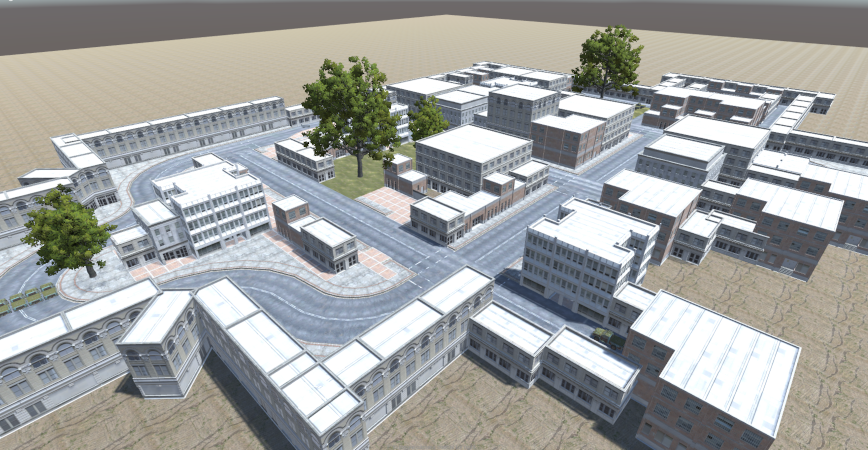}
    \includegraphics[height=.22\linewidth]{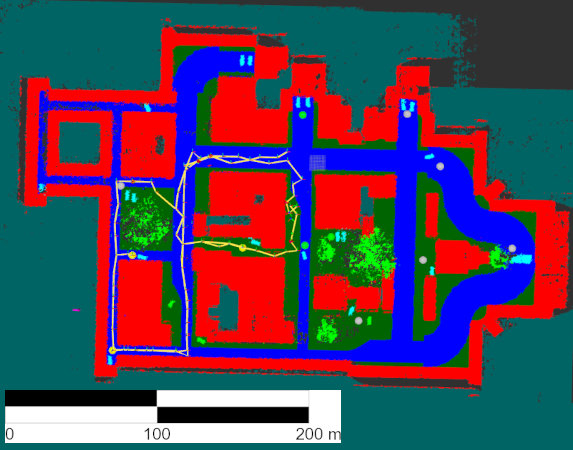}
    \caption{\textbf{\uline{Top}}: Real-world environments used for experiments. \uline{Left}: urban environment showcasing tall buildings and cars. \uline{Center}: semi-urban environment with open spaces and grass fields. \uline{Right}: rural environment with dense forest. The areas for these environments are $62500\,\text{m}^2$, $105000\,\text{m}^w$ and $157000\,\text{m}^2$ respectively.
    \textbf{\uline{Bottom}}: Environment for simulation experiments. \uline{Left}: \fer{Unity-based simulation overview, with a total area of $62570\,\text{m}^2$.} \uline{Right}: Bird's-eye view of the semantic map captured by the UAV in the simulator. }
    \label{fig:environments}
    %
    %
    %
    %
    \vspace{-.5cm}
\end{figure*}

\subsection{Real-world deployments}
We deployed the system described in Sec.~\ref{sec:methods} on our autonomous platforms, displayed in Fig.~\ref{fig:fig1}:
\begin{itemize}
    \item Ground robots: Clearpath Jackals, fitted with AMD Ryzen 3600 CPUs and an NVIDIA GTX 1650 GPU. These robots perform navigation and obstacle avoidance using an Ouster OS1-64 LiDAR.
    \item Aerial robots: our Falcon 4 \gls{uav}~\cite{liu2022large} was used for high altitude operations. This robot is equipped with a U-Blox F9P GPS, a downward-facing OVC camera~\cite{quigley2019OVC}, and a PX4-based flight controller. This system executed the communication-aware exploration algorithm described in Sec.~\ref{subsec:comm-aware-expl}.
\end{itemize}

We tested both Rajant \glspl{manet} as well as ad-hoc WiFi with Babel~\cite{chroboczek2011babel} for the \gls{phy}, both working at 2.4 GHz. Experimentally, we observed that the Rajant radios perform better than WiFi+Babel, particularly since routing tables were created faster and the association time was lower. To decrease network congestion, we did not use the mesh capabilities of the Rajant nodes and only performed one-hop communication. When the \gls{rssi} threshold was greater than 20 dBm, communication was triggered between nodes.

We performed tests in three locations, shown in Fig.~\ref{fig:environments}.  These locations showcase distinct communication conditions, from open spaces to dense urban environments. \fer{Experiments ran for 116 accumulated minutes, and the ground robots traveled an accumulated distance of 17.8 km.}
The real-time bandwidth evolution during two experiments in these locations is shown in Fig.~\ref{fig:real-world-bw}. We observe that the bandwidth spikes sporadically only when robots are within communication range. Compared to the literature~\cite{saboia2022achord}, \papertitle achieves similar bandwidth rates with far sparser communication, as the system can support extended periods of isolation between nodes. 99.5\% of the 1720 interactions in these experiments finished successfully from header exchange to transmission end, and only seven interactions were interrupted. 
This can be explained by the high \gls{rssi} chosen to trigger communications.
More importantly, we do not observe any clogging of the network when robots return to communication range.

\subsection{Simulation Experiments}
We performed simulation experiments with the twofold goal of 1) analyze the behavior of \papertitle over varying team configurations (i.e., number of UAVs and UGVs) and communication ranges and 2) stress the capabilities of \papertitle over congested communication regimes. 

\fer{We used real-world data from our experiments to design a transmission model between robots which we implement in a \todo{photorealistic, physics-based simulator in Unity with ROS interfaces} (Fig. \ref{fig:environments}). \zac{Furthermore, our simulated robots used the same perception and planning stack as our physical robots, which emulated realistic latencies and noise in our system.} We limited the total experiment time to 30 minutes.}  

Robots performed the same target search task described in Section \ref{subsec:comm-aware-expl}. 
For configurations with multiple UAVs, the second UAV only performed a SEARCH mission.


\subsubsection{Transmission Model}
We implemented a transmission model parameterized by exponential distributions describing latency per byte transmitted
\begin{align}
f(x;\lambda) = \begin{cases}
\lambda  e^{ - \lambda x} & x \ge 0, \\
0 & x < 0.
\end{cases}
\end{align}
where $\lambda$ is related to the rate of the channel.
We fitted three different distributions for the different stages of the communication: One for the header request ($\lambda_h$), one for each piece of data transmitted ($\lambda_t$), and one for the end of communication message ($\lambda_e$). These distributions aim to model the communication exchange latency as a Poisson point process.
Experimentally, we obtained: $\lambda_h=2.67$, $\lambda_e=0.51$, and $\lambda_t = 7.63$.

We imposed the constraint that only a single robot can broadcast on the wireless channel at a time to simulate network contention. This is a worst-case scenario, as real-world experiments exhibit spatial multiplexing when node cliques are well separated.

\subsubsection{Experimental Results}
Tab.~\ref{tab:results_sim} summarizes our simulation configurations and results. We limited the communication radius to mimic different \gls{rssi} thresholds that trigger communication.
We evaluated two performance metrics: the number of goals found, and the mean time to goal. We also evaluated the average message wait time and standard deviation, as a measure of channel contention.

\fer{
The results indicate that task performance is primarily driven by the number of \glspl{ugv} and channel contention. As expected, the number of \zac{found} goals increases with the number of \glspl{ugv}. Across fixed communication radii (or for fixed transmission power for the radios), the 1 \gls{uav} / 10 \glspl{ugv} teams consistently outperform other configurations in terms of the number of found goals. 
Moreover, for a single \gls{uav} increasing the radius of the communication also has a positive impact. 
\zac{Teams with 3 and 5 UGVs generally benefit from an additional UAV that acts as a data mule. However, when operating with 10 UGVs or with a 30 m communication radius, the benefit of} additional connectivity \zac{is offset by increased contention, which leads to decreased task performance}. 

The average time to goal, on the other hand, improves with larger teams and larger communication radii. Target disambiguation is an important factor for time to goal, and disambiguation messages \zac{are less} affected by contention as they have a high priority and a small size. Moreover, larger communication radii and an additional UAV helped the ground robots acquire maps and make initial plans more quickly, further improving time to goal. 
}

%% file: media/sm.tex
\begin{tikzpicture}[->,>=stealth',shorten >=1pt,auto,node distance=2.8cm,
                    semithick]
  \tikzstyle{every state}=[fill=red,draw=none,text=white, minimum size=1.3cm]
\scriptsize
  \node[initial,state] (A)              {INIT};
  \node[state]         (B) [right of=A] {SEARCH};
  \node[state]         (D) [right of=B] {COMM};
  \node[state]         (C) [below of=B] {EXPL};

  \path (A) edge              node {$t > t_i$} (B)
        (B) edge              node {$\text{found} = 1$} (D)
        (D) edge [bend left=40]  node [text width=2cm,align=left]{comm\_end = 1 .OR. $t>t_c$} (C)
        (B) edge [bend left=30]  node [text width=2cm,align=left]{$\text{found} = 0$ .AND. $t > t_s$} (C)
        (C) edge [bend left=30]  node {$t > t_e$} (B);
\end{tikzpicture}

%% file: tables/simresults.tex
\newcolumntype{N}{>{\centering\arraybackslash}m{.25in}}
\newcolumntype{G}{>{\centering\arraybackslash}m{.4in}}
\newcolumntype{T}{>{\centering\arraybackslash}m{.2in}}

\begin{tabular}{T T T T T T T }\toprule
\multicolumn{1}{N }{} & \multicolumn{2}{c}{\textbf{Configuration}} & \multicolumn{2}{c }{\textbf{Performance}} &  \multicolumn{2}{c }{\textbf{Message wait}}\\  
\cmidrule(lr){2-3}
\cmidrule(ll){4-5}
\cmidrule(ll){6-7}
\multicolumn{1}{ G }{\textbf{Comm Radius [m]}} & \textbf{UAVs} & \textbf{UGVs} & \textbf{Goals [\%]} & \textbf{Mean time [min]} & \textbf{Mean [s]} & \textbf{SD [s]} \\ 
\cmidrule(lr){1-1} \cmidrule(lr){2-3} \cmidrule(ll){4-5} \cmidrule(ll){6-7}
\multicolumn{1}{G}{5 } & 1 & 3 & 36.70\% & 23 & 0.17 & 0.87 \\
\multicolumn{1}{G}{5 } & 1 & 5 & 70.00\% & 16.4 & 0.18 & 1.29 \\
\multicolumn{1}{G}{5 } & 1 & 10 & \textbf{76.70\%} & 15.8 & 0.53 & 5.13 \\
\multicolumn{1}{G}{5 } & 2 & 3 & 40.00\% & 19.7 & 0.23 & 1.44 \\
\multicolumn{1}{G}{5 } & 2 & 5 & 73.30\% & 18.6 & 0.24 & 2.42 \\
\multicolumn{1}{G}{5 } & 2 & 10 & 70.00\% & \textbf{15.2} & 0.63 & 6.90 \\
\cmidrule(lr){1-1} \cmidrule(lr){2-3} \cmidrule(ll){4-5} \cmidrule(ll){6-7}
\multicolumn{1}{G}{10} & 1 & 3 & 46.70\% & 16.5 & 0.32 & 1.80 \\
\multicolumn{1}{G}{10} & 1 & 5 & 55.60\% & 14.9 & 0.40 & 2.85 \\
\multicolumn{1}{G}{10} & 1 & 10 & \textbf{76.70\%} & 14 & 1.33 & 13.89 \\
\multicolumn{1}{G}{10} & 2 & 3 & 53.30\% & 14.1 & 0.61 & 3.27 \\
\multicolumn{1}{G}{10} & 2 & 5 & 70.00\% & 13.8 & 0.62 & 4.93 \\
\multicolumn{1}{G}{10} & 2 & 10 & 73.30\% & \textbf{11.7} & 1.11 & 13.95 \\
\cmidrule(lr){1-1} \cmidrule(lr){2-3} \cmidrule(ll){4-5} \cmidrule(ll){6-7}
\multicolumn{1}{G}{20} & 1 & 3 & 50.00\% & 14.6 & 0.38 & 2.66 \\
\multicolumn{1}{G}{20} & 1 & 5 & 55.60\% & 13.7 & 0.49 & 3.76 \\
\multicolumn{1}{G}{20} & 1 & 10 & \textbf{86.70\%} & 14 & 1.47 & 25.73 \\
\multicolumn{1}{G}{20} & 2 & 3 & 43.30\% & 16 & 1.05 & 6.02 \\
\multicolumn{1}{G}{20} & 2 & 5 & 66.70\% & \textbf{12.1} & 0.95 & 8.43 \\
\multicolumn{1}{G}{20} & 2 & 10 & 75.60\% & 13.8 & 3.91 & 55.39 \\
\cmidrule(lr){1-1} \cmidrule(lr){2-3} \cmidrule(ll){4-5} \cmidrule(ll){6-7}
\multicolumn{1}{G}{30} & 1 & 3 & 40.00\% & 16.1 & 0.36 & 2.62 \\
\multicolumn{1}{G}{30} & 1 & 5 & 66.70\% & 14.9 & 0.95 & 6.85 \\
\multicolumn{1}{G}{30} & 1 & 10 & \textbf{93.30\%} & 13.9 & 3.21 & 35.45 \\
\multicolumn{1}{G}{30} & 2 & 3 & 60.00\% & 18.3 & 1.20 & 6.60 \\
\multicolumn{1}{G}{30} & 2 & 5 & 62.20\% & 12 & 1.50 & 12.08 \\
\multicolumn{1}{G}{30} & 2 & 10 & 73.30\% & \textbf{10.5} & 6.80 & 80.92 \\
\bottomrule
\end{tabular}\textbf{}

%% file: sections/conclusion.tex
\section{Conclusion}
In this work, we proposed \papertitle: a resilient, gossip-based opportunistic communication and coordination framework for multi-robot field deployments. \papertitle enables large-scale collaboration between teams of heterogeneous robots, without imposing limitations on disconnection time, and provides network and communications metrics that can be exploited to inform higher-level autonomy behaviors. 
\fer{We also presented an example of communication-aware exploration by an aerial vehicle.}

\fer{We performed extensive testing of \papertitle in simulation and real-world experiments. We explored the limitations of our method as the number of agents increased, and we showed that network contention does not affect performance significantly in our 1 \gls{uav} / 10 \glspl{ugv} simulation. Our system scales well when both the size of the environment and the number of robots grow. A higher number of nodes may reduce network performance due to contention, particularly if the communication radius is comparable to the size of the environment. In these situations ensured network communication schemes may be preferable. 
}

Incorporating network models from the environment based on the semantic map obtained from the \gls{uav} is an appealing path for improvement. For instance, ground robots could plan routes towards dynamic rendezvous points, maximizing communication probability with other nodes in the system.